# Exploration of Reproducible Generated Image Detection

Yihang Duan

## Abstract

While the technology for detecting AI-Generated Content (AIGC) images has advanced rapidly, the field still faces two core issues: poor reproducibility and insufficient generalizability, which hinder the practical application of such technologies. This study addresses these challenges by reviewing 7 key papers on AIGC detection, constructing a lightweight test dataset, and reproducing a representative detection method. Through this process, we identify the root causes of the reproducibility dilemma in the field: firstly, papers often omit implicit details such as preprocessing steps and parameter settings; secondly, most detection methods overfit to exclusive features of specific generators rather than learning universal intrinsic features of AIGC images. Experimental results show that basic performance can be reproduced when strictly following the core procedures described in the original papers. However, detection performance drops sharply when preprocessing disrupts key features or when testing across different generators. This research provides empirical evidence for improving the reproducibility of AIGC detection technologies and offers reference directions for researchers to disclose experimental details more comprehensively and verify the generalizability of their proposed methods.

## Introduction

In recent years, Generative Artificial Intelligence (AIGC) technology has developed rapidly and been widely applied, significantly lowering the threshold for generating high-fidelity images using AI tools. With the continuous iteration of model architectures—from early Convolutional Neural Networks (CNNs) and Generative Adversarial Networks (GANs), to current mainstream Diffusion models, and further to the latest autoregressive and FlowMatch architectures—the fidelity of generated images has reached a level of being "indistinguishable to the naked eye." While this technological advancement has injected efficiency into fields such as creative design and content production, it also provides opportunities for malicious activities: for instance, using generated images to spread fake news, creating infringing works that mimic the style of specific artists, and even forging identity information to commit fraud. Consequently, the necessity and urgency of "AIGC image detection" technology have become increasingly prominent, and related detection algorithms have since emerged as a research focus in both academia and industry.

The core challenge in current AIGC image detection stems from the "method lag" caused by the rapid iteration of generative models. Early detection methods targeting GAN-generated images have been proven ineffective for content generated by Diffusion models; and as Diffusion models have become the primary source of current AIGC images due to the popularity of commercial products such as Stable Diffusion (Stability AI (2025) ) and MidJourney(MidJourney Inc. (2025) ), a large number of detection methods specifically designed for Diffusion-generated images have emerged in academia. Most of these methods claim to possess "cross-generator generalization ability" in their papers—meaning they can effectively detect images produced by new generative models not seen during training—but they have exposed numerous issues in practical applications, making it difficult to achieve real-world deployment.

From a technical perspective, AIGC image detection is essentially a binary classification task between "real images" and "generated images," and existing methods can be categorized into three types: The first type continues the traditional GAN detection approach, focusing on local image differences (e.g., high-frequency noise, texture deviations), and enhances the extraction of such differential features by designing dedicated modules (e.g., attention mechanisms, multi-scale feature fusion modules) to distinguish between real and generated images; The second type designs training-free detection schemes based on the inherent characteristics of generative models, with the most representative being "reconstruction error-based methods"—which achieve classification by leveraging differences in reconstruction errors between real and generated images (Chu et al. (2024);Ricker et al. (2024); Wang et al. (2023) )—but such methods generally have interpretability flaws, and no existing research has elaborated on the underlying principles of reconstruction errors in detail; The third type explores cross-modal fusion approaches, such as introducing Large Language Models (LLMs) to analyze image semantics and generation logic for improving the interpretability of detection (Zhang et al. (2025) ), but they have not yet been applied on a large scale due to limitations in detection efficiency.

Despite the widespread attention on the claimed performance of various detection methods, their "reproducibility" and "practical applicability" have significant shortcomings, which severely hinder technical deployment. We investigated 7 AIGC image detection papers published in top conferences in the computer science/AI field, and after organizing their code repositories and community feedback, we found that the problems in the reproduction process are specifically manifested in three aspects: Poor code availability, as some papers do not make their code public or only provide fragmented scripts; Discrepancies between reproduced results and those reported in the papers, mainly due to insufficient specific reproduction details; Significant performance degradation when methods are transferred to new datasets—a phenomenon that raises great doubts about the effectiveness of detection methods (see Figure 1). In-depth analysis shows that the core cause of this problem is likely that such methods are only trained on "images generated by a single generator," leading to the features learned by the model lacking generality and being unable to adapt to the practical needs of multi-scenario and multi-generator applications. Previous research has pointed out this possibility, and this paper further verifies the validity of this view in some scenarios by reproducing that work and expanding experiments.

| CNNSpot | FreDect | Fusing | GramNet | LNP |
|---|---|---|---|---|
| 60.89 | 57.22 | 57.09 | 60.95 | 58.52 |
| 9.86/99.25 | 0.89/99.55 | 0.02/99.98 | 4.76/99.66 | 7.72/96.70 |

Figure 1: Studies (Zhang et al. (2025) ) show that some methods perform poorly on datasets in real-world application scenarios (Yan et al. (2024) ). The data on the left and right sides represent the detection accuracy for real and fake images.

To address the aforementioned key issues of "difficult reproduction and poor generalization," as well as the potential causes of poor generalization, this paper conducts an exploratory study through empirical research, with core contributions as follows:

- By investigating 7 AIGC image detection papers, and combining community discussions with independent reproduction, we systematically analyze the core factors affecting the reproducibility of methods in this field;
- We construct a lightweight test dataset that can quickly verify the generalization and reproducibility of detection methods; at the same time, we point out that a comprehensive and effective test set needs to cover multiple types of generators and real social media data to more accurately evaluate the practical application potential of methods;
- We reproduce a representative detection method (Rajan et al. (2024) ) and conduct in-depth analysis of its principles, extract key conclusions explaining the limitations of detection methods, and provide directions for subsequent method improvements.

## Survey Results:

To systematically sort out the reproducibility issues in the field of AIGC image detection, we selected 7 representative relevant papers (Chu et al. (2024); Cozzolino et al. (2024); Rajan et al. (2024);Ricker et al. (2024); Tan et al. (2024);Wang et al. (2023); Zhang et al. (2025) ) as research objects. These papers were all published in top computer science/AI conferences such as AAAI, covering the mainstream technical routes of AIGC detection mentioned earlier, with a time span from 2023 to 2025, which can provide valuable references for analyzing reproducibility issues in the field. The survey dimensions include the original papers, community discussions (e.g., GitHub Issues, comments on Papers With Code), and descriptions of performance reproduction in subsequent studies that cite these papers. The core is to infer the key underlying reasons through "phenomena of reproduction failure/deviation."

### 2.1 Code Availability:

Current mainstream AI conferences all recommend that submitted papers open-source their code to lower the threshold for reproduction, but the survey found that the field of AIGC image detection still has issues of "unopen-sourced code" or "incomplete open-sourcing." We classified them according to two criteria: "whether code access is open" and "whether directly runnable detection tools (e.g., pre-trained models, detection code) are provided," with the results as follows:

Completely unopen-sourced: Only 1 out of the 7 papers did not open-source any code (Cozzolino et al. (2024) ), with doubts in its community discussion area about "inability to verify the effectiveness of the method," and the authors did not respond to "whether they plan to open-source;"

Incomplete open-sourcing: 1 paper only provided training code (Tan et al. (2024) ), requiring separate model training to verify the effect, which hinders reproducible research;

Complete and convenient open-sourcing: The remaining 5 papers all made public directly usable detection methods (including complete code, with some providing pre-trained models), and community feedback during the reproduction process was smoother.

It should be particularly noted that some papers have the situation of "complementing code over time"—initially only partial code was made public, and it was gradually supplemented completely after community inquiries. Therefore, when investigating the impact of code availability on reproducibility, it is necessary to clarify the "survey time node" to avoid judgment bias caused by "supplementary open-sourcing."

### 2.2 Reproduction Performance Deviation:

Among the 7 surveyed papers, 4 had cases where "reproduced results were inconsistent with the detailed performance data reported in the original papers." For 3 of these, the causes were clarified or solutions were found through community discussions with the original authors:

For example, in the code repository of a certain paper (Rajan et al. (2024) ), community members pointed out that "reproduced results were inconsistent with the original paper." After multiple rounds of discussion, it was confirmed that the discrepancy stemmed from "dataset preprocessing operations different with the original paper," and the reproduced results became consistent with the original paper after supplementing these operations.

In addition, potential reasons for reproduced performance being inferior to the original paper include: missing key functional modules in the code, or critical flaws in the code itself—such issues can directly halt reproduction work, requiring community members to troubleshoot and correct them on their own, significantly increasing reproduction costs. Only the reproduction deviation of 1 paper was not resolved through community discussions.

**2.3 Cross-Dataset Generalization**:

Most detection methods perform well on self-built datasets in the original papers and claim to "effectively generalize to unseen datasets," but in actual cross-dataset tests, their performance generally degrades significantly, with specific phenomena as follows: As shown in Figure 1, multiple mainstream detection methods performed far below the level reported in the original papers on a certain dataset, and even produced "completely wrong judgments," raising doubts about "the actual effectiveness of these methods;"

A detection method (Wang et al. (2023) ) exhibits significant performance variations across different datasets. For example, the accuracy drops from 100% to 47% (Chu et al. (2024) ), or from 99% to 50% (Zhang et al. (2025) ), casting doubt on the universality of its core principles.

Based on survey analysis, the core reasons for poor generalization performance include two points: Different generators (even with similar technical routes) have inherent differences in structure and generation logic, and detectors trained only on "data from a single generator" struggle to adapt to multi-generator scenarios; Interfering factors during training (such as feature damage caused by differences in dataset preprocessing operations) further weaken the method's cross-dataset adaptability, increasing the difficulty of reproducing consistent performance.

**2.4 Summary of Reproducibility Issues**

Based on the comprehensive survey results, the current reproducibility issues in AIGC image detection papers can be summarized into the three core obstacles mentioned above, and these three types of issues are interrelated, collectively restricting technical deployment: incomplete or unopensourced code directly hinders reproduction, while reproduction performance deviations and poor cross-dataset generalization further weaken the credibility of methods—making it even difficult to determine whether "poor performance is due to code issues or the method itself." These issues not only hinder the inheritance and iteration of research results but also plunge the field into the dilemma of "a large number of methods emerging without reliable verification benchmarks," which is the core direction that this paper attempts to address through "dataset construction and empirical reproduction" in the following sections.

## Construction of the Test Dataset

Based on the issues identified earlier, to conveniently and quickly measure the reproducibility of an AIGC image detection method, a small yet comprehensive test dataset is needed. We believe that the effectiveness and flaws of methods can be verified using such a comprehensive dataset, and the reproducibility of methods and the interpretability of influencing factors can be theoretically clarified. Previous datasets or some commonly used standard datasets have problems such as overly outdated generators and limited types of generators; conclusions drawn from such datasets obviously cannot be reliably generalized to the latest generators or real-world applications. Therefore, we constructed a small but relatively comprehensive dataset to rapidly study the effectiveness and generalizability of a method, and to attempt to explain its strengths and limitations. The composition of the dataset is as follows: 1k real images and 1k generated images from chameleon (Yan et al. (2024) ) (using the uniform sampling method; data obtained from other datasets below are also obtained using the uniform sampling method), 1k images each from three generators (sdv2, sdxl, flux), 1k real images, 1k images crawled from the AI creation community civitai (CivitAI (2025) ) using a web crawler, and 1k images crawled from human artists' works (Pixiv Inc. (2025) ) using a web crawler. Next, we will attempt to reproduce and study a paper using this dataset.

## Reproduction Experiment

We selected the paper "Aligned Datasets Improve Detection of Latent Diffusion-Generated Images" (published in ICLR 2025) (Rajan et al. (2024) ) as the core object for reproduction. The detection idea proposed in this paper is concise and representative, and its publicly available complete code and pre-trained models provide a basis for reproduction; meanwhile, through reproduction and extended experiments, we not only verified the effectiveness of this method but also revealed the key reasons for the poor generalization and difficult reproduction of AIGC detection methods.

This paper proposes a simple idea: use VAE to reconstruct real images as generated images, then feed the real images and their reconstructions into a simple classifier for training. This simple data alignment method achieves excellent results and has stronger anti-interference ability compared to models trained directly using real images and other generated images. Even randomly generated images and their reconstructions can be used to train models to recognize the features of such generators. The reconstruction methods used in this paper are VQ-VAE and SD3's VAE. During the survey, this paper made all code publicly available in its code repository, along with the models trained on the dataset used in the paper. This enables rapid reproduction and testing. There was an issue in the code repository regarding discrepancies between the reproduced results and the results reported in the paper on the original paper's dataset. After discussions with the authors, it was found that the discrepancies were caused by changes in the settings of some parameters and differences in certain preprocessing steps during reproduction. After making some modifications, the results became significantly closer to those reported in the paper. Next, we will attempt to reproduce this paper on our dataset, test the performance of the pre-trained models, and explain some phenomena observed during reproduction through additional experiments. These conclusions can also be used to explain why other papers show poor performance during reproduction.

Our reproduction is divided into two parts: training and testing. In the training part, we follow the insights proposed in this paper; in the testing part, we test the performance of the pre-trained models and the models we trained ourselves on our dataset. It should be noted that we strive to use the code open-sourced by the authors for the entire process, making modifications only where necessary—such as changing the multi-threaded operation of the training program to single-threaded (i.e., training one model at a time) to adapt to our hardware equipment. In other words, we strive to keep everything consistent with the original experiment except for the hardware equipment, environment, and necessary changes made to adapt to these environments. The only modification made is to the dataset, which is intended to verify the true generalization of the theory proposed in this work.

| Train Data\Test Data | AI | | | | | NATURE | | |
|---|---|---|---|---|---|---|---|---|
| | sd | sdxl | flux | chame | civitai | chame | pixiv | real |
| sd | 97.3 | 16.8 | 35.3 | 22.1 | 1.9 | 94.5 | 96.1 | 92.1 |
| sdxl | 92.8 | 61.0 | 62.5 | 2.3 | 0.33 | 98.6 | 99.6 | 94.4 |
| flux | 57.9 | 0.1 | 98.3 | 0.8 | 0.01 | 99.1 | 100 | 99.6 |
| chame | 47.7 | 82.3 | 66.4 | 89.1 | 85.4 | 69.8 | 24.9 | 47.5 |
| sync | 99.9 | 39.8 | 25.6 | 28.8 | 18.4 | 96.7 | 99.2 | 87.9 |

Figure 2: This refers to the performance of three types of models: models trained on reconstructions from the VAEs of different generators (based on the method described in the original paper), models trained on the Chameleon dataset (Yan et al. (2024) ), and models trained in the original paper.

On our dataset, we trained a model on the reconstructions corresponding to each generator; in addition, we trained a model on the chameleon (Yan et al. (2024) ) dataset. For each model we trained ourselves, we used the results from the 10th epoch for testing. Figure 2 presents the test results of the models we trained. From these results, a clear conclusion can be drawn: models trained on reconstructions from a specific VAE can effectively detect images generated by models with that VAE, but struggle to generalize to other generators. This undoubtedly poses challenges for applications in real-world scenarios. However, it is also worth noting that this indicates that the reconstructions or generations of each VAE have their unique features. Next, we will analyze the features of images generated by these VAEs in detail.

Through direct visual observation, it can be found that there are no differences in content or semantics between real images and their reconstructions. Therefore, based on previous knowledge, it is natural to hypothesize that these features belong to high-frequency information. We thus conducted frequency domain analysis on these images.

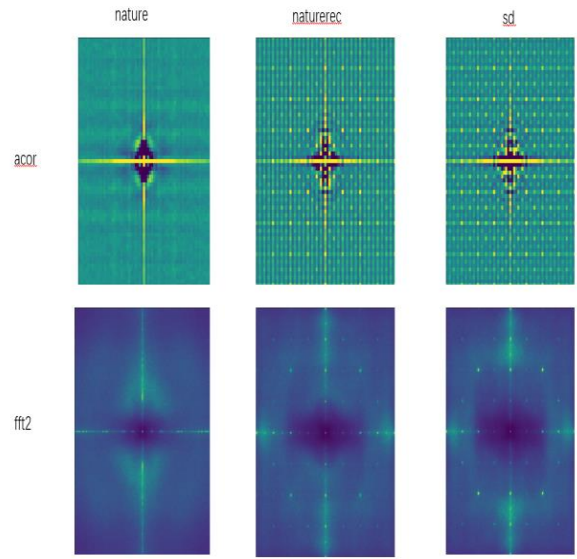

Figure 3: This is our frequency domain and energy analysis of real images, reconstructions of real images, and corresponding generator-generated images. The upper part shows the energy spectra, and the lower part shows the frequency spectra. It can be observed that real images after reconstruction and generator-generated images have similar frequency and energy characteristics.

Figure 3 illustrates the following conclusion: after real images are reconstructed by the VAE of the corresponding generator, they acquire the same high-frequency features as those of the generator; moreover, the features of different VAEs have certain differences. This also explains why detectors trained on one type of generator struggle to detect various unseen data.

Now that we can confirm that the features used for model classification exist in the high-frequency part, it is well-known that information in the high-frequency part is prone to being affected or lost during various processing steps. Our subsequent experiments will investigate the impact of processing methods on these high-frequency features.

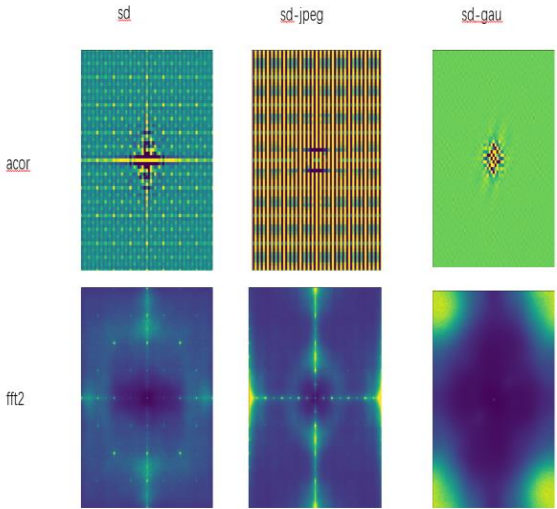

Figure 4: This is the analysis of the energy spectra and frequency spectra of generator-generated images and their versions after JPEG compression and Gaussian noise addition. It can be observed that these operations cause the inherent features of the generators themselves to be obscured.

Figure 4 shows the changes and analysis of these high-frequency features after subjecting real images reconstructed by SDv2's VAE and images generated by SDv2 to common processing steps (JPEG compression, resizing operations, and Gaussian noise addition). It can be seen that the original features of SDv2's VAE have completely disappeared. This implies that if the training data undergoes preprocessing, it is highly likely to impair the performance of the trained detector—and this is also the reason for the discrepancies between the reproduced results and the reported results in some work (Chu et al. (2024) ).
To confirm that preprocessing operations impair the performance of trained detectors, we tested two types of models: one trained using real images and their reconstructions, and the other trained using JPEG-compressed real images and their previous reconstructions. The test results are shown in Figure 5.

|   | na | najpeg | ai | aijpeg |
|---|---|---|---|---|
| sd | 92.1 | 97.5 | 99.3 | 3.2 |
| A | 26.6 | 83.1 | 100 | 21 |

Figure 5: This is a comparison of the performance between models trained on normal data and Model A trained on JPEG-compressed data. It can be observed that Model A completely takes JPEG compression artifacts as the classification criterion.

The experimental results show that JPEG compression does destroy the original features and causes the model to recognize these damaged features as the characteristics of a certain type of image. Based on prior knowledge, JPEG compression does produce high-frequency artifacts, and resizing and Gaussian noise addition also inevitably introduce high-frequency artifacts. It is reasonable to believe that these operations will also impair detector performance. This also implies that training simple classifier models may lead the models to only learn to classify by identifying the presence of a specific feature; during actual detection applications, the classifier will simply make classifications based on whether this feature exists. This explains the specific phenomena observed for various trained detectors on a single dataset in Figure 1. This reminds subsequent detector designers to pay attention to dataset balance to avoid models over-relying on a specific feature to make decisions.

## Discussion

Experimental results show that generators with different structures have unique high-frequency features. If an AIGC image detector is only trained on a specific dataset, it tends to overfit to that type of generator, which in turn leads to a significant decline in its reproducibility in other scenarios. Based on this, a key question arises: Do there exist common features among different generators that can be used to simplify model training and detection processes?

In response to this question, our in-depth research on the work (Rajan et al. (2024) ) shows that the "VAE-specific features" learned by the model through VAE reconstruction essentially stem from the structural design of VAE, and more fundamentally, from the inherent characteristics of VAE's convolutional layers. In Diffusion models, VAE has two core roles: first, providing the latent space required for the diffusion process to reduce computational costs; second, completing image dimension elevation in the final step of image generation—it is precisely this dimension elevation step that injects the VAE-specific artifacts into the image. It is worth noting that early VAEs, due to their fixed-step convolutional design, caused periodic artifacts in generated images; however, with the iterative optimization of generative model structures, such easily detectable explicit artifacts are gradually decreasing, further increasing the difficulty of cross-generator detection.

In addition, a recent study (Heo and Woo (2025) ) points out that in multi-step generation processes, the finally retained artifact traces mainly come from the last step of the generation process. This conclusion precisely explains "why VAE reconstruction-based detection methods can work efficiently"—as the last link in the generation process of Diffusion models, the artifacts introduced by VAE are key identifiers of generated images, and the model can achieve detection by learning this feature without capturing features from other steps.

Integrating the above conclusions, when reproducing methods related to AIGC image detection, if the results are not as expected, we can look for reasons from the perspective of "feature essence and generation process": for example, whether the reproduction performance deviation stems from "the specific high-frequency features relied on by the detection method being damaged due to generator iteration or preprocessing operations"; or whether it is because "the method only learned the exclusive artifacts of a certain type of generator and cannot adapt to the feature distribution of other generators". This analytical approach can provide a clear direction for problem localization in subsequent reproduction work and also offer theoretical references for designing detection methods with better generalization.